\documentclass[conference]{IEEEtran}
\IEEEoverridecommandlockouts
\usepackage{cite}
\usepackage{amsmath,amssymb,amsfonts}
\usepackage{algorithmic}
\usepackage{graphicx}
\usepackage{textcomp}
\usepackage{xcolor}
\usepackage{hyperref}
\usepackage{xspace}
\usepackage{booktabs}
\usepackage{multirow}
\usepackage{algorithmic}
\usepackage[linesnumbered,ruled,vlined]{algorithm2e}
\def\BibTeX{{\rm B\kern-.05em{\sc i\kern-.025em b}\kern-.08em
    T\kern-.1667em\lower.7ex\hbox{E}\kern-.125emX}}

\newcommand{\Span}{\Psi} 
\newcommand{\gyz}[1]{#1}
\newcommand{\gy}[1]{#1}

\newcommand{\pk}[1]{#1}
\newcommand{\name}{\textsc{Shrink}\xspace} 
\newcommand{\simpiece}{\textsc{SimPiece}\xspace} 
\renewcommand{\footnoterule}{
    \kern -3pt
    \hrule width 0.95\columnwidth
    \kern 2.6pt
}
\begin{document}

\title{
    Highly Efficient Direct Analytics on Semantic-aware Time Series Data Compression
}

\author{
\IEEEauthorblockN{Guoyou Sun\textsuperscript{a}, Panagiotis Karras\textsuperscript{b,c}, Qi Zhang\textsuperscript{a}}
\IEEEauthorblockA{
 \textit{\textsuperscript{a} DIGIT and Department of Electrical and Computer Engineering, Aarhus University, Denmark} \\
 \textit{\textsuperscript{b} DIGIT and Department of Computer Science, Aarhus University, Denmark} \\
 \textit{\textsuperscript{c} Department of Computer Science, University of Copenhagen, Denmark} \\
Email: \{guoyous, qz\}@ece.au.dk,  piekarras@gmail.com
}
}

\maketitle

\begin{abstract}
Semantic communication has emerged as a promising paradigm to tackle the challenges of massive growing data traffic and sustainable data communication. It shifts the focus from data fidelity to goal-oriented or task-oriented semantic transmission. While deep learning-based methods are commonly used for semantic encoding and decoding, they struggle with the sequential nature of time series data and high computation cost, particularly in resource-constrained IoT environments. Data compression plays a crucial role in reducing transmission and storage costs, yet traditional data compression methods fall short of the demands of goal-oriented communication systems. \pk{In this paper, we propose a novel method for direct analytics on time series data compressed by the \name compression algorithm. Through experimentation using outlier detection as a case study, we show that our method outperforms baselines running on uncompressed data in multiple cases, with merely 1\% difference in the worst case. Additionally, it achieves four times lower runtime on average and accesses approximately 10\% of the data volume, which enables edge analytics with limited storage and computation power. These results demonstrate that our approach offers reliable, high-speed outlier detection analytics for diverse IoT applications while extracting semantics from time-series data, achieving high compression, and reducing data transmission.}
\end{abstract}
\section{Introduction}

Semantic communication has recently emerged as a novel communication paradigm to address the challenges of ever-increasing diverse data traffic, which focuses on transmitting task-relevant information, referred to as \emph{Semantics}, instead of reliably transmitting sequence of bits~\cite{xie2020lite}. Existing works predominantly use deep learning-based methods for semantic extraction, encoding and decoding~\cite{zhang2022deep}. However, deployig these approaches in resource-constrained IoT devices is challenging due to high computational costs of training and inference~\cite{kutay2024}. Besides, most of these methods were developed for computer vision and natural language processing, where inductive biases like cropping-invariance and transformation-invariance are prevalent~\cite{yang2022semantic}. While these biases are effective for images, they fail to generalize to time series data~\cite{yue2022ts2vec}. 

\begin{figure}[!ht]
\vspace{-1mm}
\centering
\includegraphics[width=\linewidth]{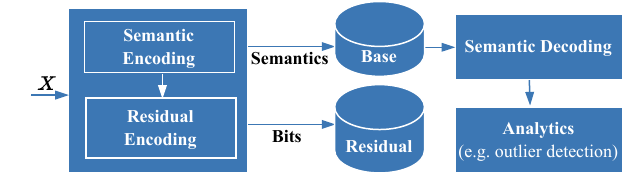}
\vspace{-5mm}
\caption{System diagram for compressed data analytics with \emph{Base}.}\label{fig:framework}
\vspace{-5mm}
\end{figure}

To overcome these limitations, model-based data compression has proven to be an effective solution for time series data, providing compact representations at low computational costs~\cite{zhang2022goal}. Recent methods, such as those utilizing piecewise linear approximation (PLA)~\cite{boursier2018plato, kitsios2023sim, yang2023most}, have achieved high compression with deterministic error bounds. However, these methods mainly focus on compression performance, giving limited consideration to direct analytics on compressed data; hence, often overlook multiscale contextual information, which is crucial for tasks requiring fine-grained representations, such as anomaly detection. Generalized Deduplication (GD)~\cite{vestergaard2020randomly} and its extensions~\cite{hurst2022glean} provide lossless compression with random access capability and demonstrate the possibility to perform clustering directly on part of compressed data. GD separates compressed data into two parts and transmits only essential information for direct analytics, which improves network transmission efficiency. However, GD is not able to capture contextual information at multiple resolutions to support tasks that need fine-grained representations.

\pk{In this paper, we show how to perform analytics tasks directly on semantics, with outlier detection as an example; we build on our recent work on semantic-aware compression, \gyz{\name~\cite{sun2024shrink}}, a lightweight PLA-based approach for semantic extraction of time series data. We extend \name by introducing semantic quantization as Semantic Encoding and transformation as Semantic Decoding for direct analytics on semantics (i.e., a compact data representation). Unlike conventional data compression methods, our method captures multiscale semantic representations by adaptively grouping data points along the temporal dimension, allowing efficient analysis of arbitrary sub-series. As~\autoref{fig:framework} shows, the compressed output consists of a \emph{Base} which retains essential long-term semantics and \emph{Residuals} that capture fine-grained details. 
\noindent We rely solely on the \emph{Base} for outlier detection, significantly reducing the transmitted data volume while preserving detection accuracy. Our main contributions are summarized as follows:}

\begin{enumerate}
    \item \pk{We propose a pipeline for data analytics on data compressed by semantic-aware compression.}
    \item \pk{Through experimentation on synthetic and real datasets, we show that our compressed data analytics achieves accuracy in outlier detection comparable to that on uncompressed data (within a 1\% drop in the worst case), while it accesses around~10\% of the data size and speeds up analytics by four times on average.}
    \item \pk{We demonstrate that our framework is applicable and effective for semantic communication within IoT ecosystems, achieving lightweight semantic extraction, encoding, and decoding for time-series data.}
\end{enumerate}
\begin{figure}[!ht]
\vspace{-1mm}
\centering
\includegraphics[width=0.9\linewidth]{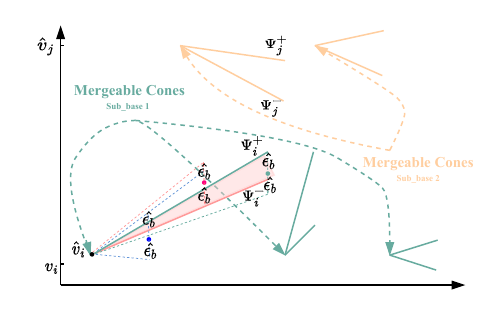}
\vspace{-4mm}
\caption{Process of Adaptive Shrinking Cone to extract semantics.}\label{fig:cone}
\vspace{-1mm}
\end{figure}

\section{BACKGROUND}

\subsection{SHRINK}

\pk{\name~\cite{sun2024shrink} is a semantic-aware compression method that achieves ultra-accurate data compression. It constructs a semantic-level \emph{Base} and encodes bit-level \emph{Residuals}, which can provide both lossless and error-bounded lossy time series data compression. \name builds a single encoding~$C_x$ that is separable into a \emph{Base}~$B_x^{\epsilon_b}$ and a \emph{Residual}~$R_x^{\epsilon}$ at various~$L_{\infty}$ error resolutions, which provide multiresolution decompression of a single encoding. To effectively capture data semantics, it employs Adaptively Shrinking Cones (ASCs) that progressively reduce their span according to data characteristics. For example, in~\autoref{fig:cone}, the span of a cone is defined by an upper slope~$\Span_i^+$ and a lower slope~$\Span_i^-$, set so that any line between them approximates the data points in the cone's segment within error threshold~$\hat{\epsilon}_b$. The data segment expands with each newly included data point, further narrowing the cone span, so that lines of slope therein approximate all data points in the segment within~$\hat{\epsilon}_b$; when the cone span becomes empty, the expansion terminates. By an adaptive base error threshold, when data values in the default interval length vary less, the cone's span grows to accommodate more data, as~\autoref{fig:fluctuation} shows. Conversely, with high data variability, the cone span reduces, due to a tighter error margin~$\hat{\epsilon}_b$. This ASC mechanism enables further efficient merging and joint representation of cones, which are transformed to linear segments~$S$ for each merged group to construct the \emph{Base}. \name encodes the remaining data details as \emph{Residuals} to support both lossy and lossless compression.}


\subsection{Outlier Detection}

\pk{We tested our direct analytics method on compressed data on three popular outlier detection baselines: IForest, DBScan, and AutoEncoder. IForest~\cite{liu2008isolation} uses a tree structure to isolate anomalies by recursive partitioning; DBScan~\cite{jin2019novel} clusters data points by density and labels unclustered points as outliers; AutoEncoder~\cite{sakurada2014anomaly} employs a neural network to learn a compressed data representation, flagging deviations as anomalies. Each baseline targets different types of outliers.}

\noindent \textbf{Point-based Outliers:} This category includes point outliers and contextual outliers. Point outliers are individual data points that significantly differ from most data, such as sudden spikes or drops in a time series. Contextual outliers are anomalies that depend on the data's context, like seasonal or trend-based deviations, and may not be globally distinct but are unusual within a context.

\noindent \textbf{Sequence-based Outliers:} \pk{This category refers to groups of data points that are normal individually but deviate collectively. For instance, a sequence of data values forming an unusual pattern can be considered sequence-based outliers.}

\begin{figure}[!h]
\vspace{-1mm}
\centering
\includegraphics[width=0.9\linewidth]{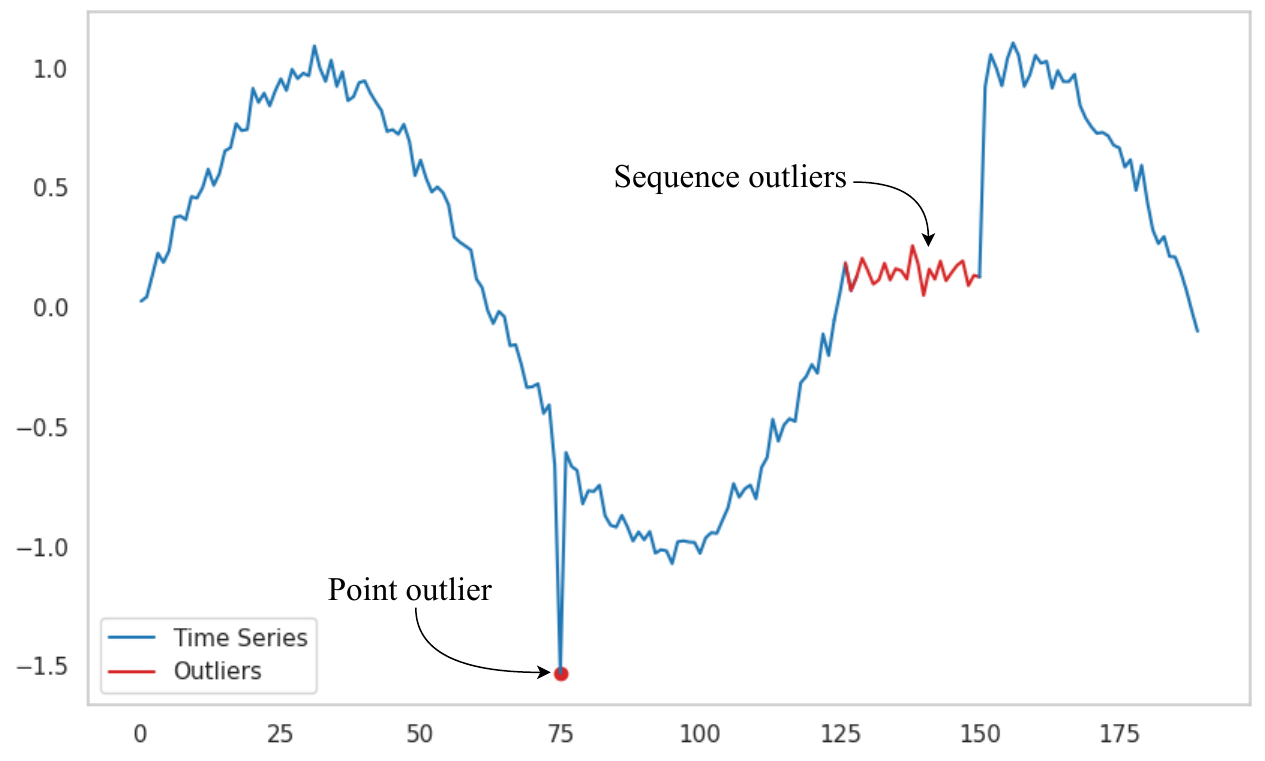}
\vspace{-3mm}
\caption{Illustration of the typical point-based and sequence-based outliers.}\label{fig:outliertype}
\vspace{-1mm}
\end{figure}

\noindent \autoref{fig:outliertype} depicts a time series with examples of both point-based and sequence-based anomalies.

\pk{The metrics, ROC AUC and PR AUC, are commonly used to assess outlier detection model performance. ROC AUC measures the model’s ability to distinguish between normal and outlier data points. It does this by plotting True Positive Rate, $\text{TPR} = \frac{\text{TP}}{\text{TP}+\text{FN}}$ against False Positive Rate, $\text{FPR} = \frac{\text{FP}}{\text{FP}+\text{TN}}$ at various anomaly thresholds, where P stands for `positive', N for `negative', T for `true', and F for `false'. A higher ROC AUC indicates that the model effectively separates anomalies from normal data across various thresholds. PR AUC captures the trade-off between precision, $\text{P} = \frac{\text{TP}}{\text{TP} + \text{FP}}$ and recall, $\text{R} = \text{TPR}$. A higher value of PR AUC indicates that the model accurately and inclusively detects anomalies.}
\section{Methodology}

In this section, we first introduce semantic quantization, a critical component of Semantic Encoding, which preserves essential semantic information. Next, we describe the transformation of compressed data, a key step in Semantic Decoding, which enables effective outlier detection.

\subsection{Semantic Quantization}\label{sec:SemQuant}

Semantic quantization is designed to preserve the underlying data features by approximating values within semantically coherent regions, effectively prioritizing significant changes over local noise. To obtain the semantics, the adaptive shrinking cone is constructed  by a dynamic error threshold, as below:

\begin{equation}\label{eq:epsilon}
\hat{\epsilon}_{b,i} = \epsilon_{b} \cdot e^{\frac23 - \beta_i}~,
\end{equation}

\noindent where $i$ is the ID of interval\footnote{For the details of how to set the length of interval, see~\cite{sun2024shrink}.} split by ASC,~$\beta_i = \frac{\Delta_i}{\Delta}$, $\Delta$ is the global value range of the whole data series, $\Delta_i$ is the local value range in interval~$i$, and~$\epsilon_b$ is the default error threshold of \emph{Base}. We can specify default quantization level with $\epsilon_{b}$ to determine the actual value $\hat{\epsilon}_{b,i}$~\cite{sun2024shrink}. The process can be implemented simply using multiplication and division operations, where the original point value  $v_i$  of shrinking cone is quantized to $\hat{v_i}$, and other points inside the cone are approximated by a linear model constructed by shrinking cone.


Notably specifying the quantization level $\epsilon_{b}$ manually is not trivial, especially without prior knowledge of the data distribution. Instead, we leverage Signal-to-Noise Ratio (SNR) to determine $\epsilon_{b}$, ensuring that the quantization process maintains a balance between compression and data fidelity~\cite{wang2022frequency}. The SNR,  can be mathematically represented as:

\begin{equation}\label{eq:snr}
\eta = 10 \log_{10} \left( \frac{\sum\limits_{i=0}^{n-1} v_i^2}{\sum\limits_{i=0}^{n-1} \left( v_i - \left\lfloor v_i \cdot 2^{-\tau} \right\rfloor \cdot 2^{\tau} \right)^2} \right)~,
\end{equation}

\noindent where $\tau$ is an integer and $2^{\tau}$ is the default quantization level. The objective of semantic quantization is to find the maximum value of $\tau$ to ensure a proper semantic quantization level. As the upper limit of quantization error $\lvert v_i - \left\lfloor v_i \cdot 2^{-\tau}\right\rfloor \cdot 2^{\tau} \rvert$ is $2^{\tau}$, we substitute it into~\eqref{eq:snr}  and set the initial value of $\tau$ as:

\begin{equation}\label{eq:quantlevel}
\tau =  \left\lfloor \frac{1}{2} \log_2  \left( {\frac{10^{-\eta / 10}}{n} \cdot \sum\limits_{i=0}^{n-1} v_i^2} \right) \right\rfloor + 1~,
\end{equation}

\noindent We initiate the search from $\tau$ and continue until  SNR falls below $\eta$. With the actual quantization level $\tau $ determined by SNR, the adaptive error threshold of \emph{Base} is then given by: 

\begin{equation}\label{eq:baseError}
\hat{\epsilon}_{b,i} = 2^{\tau} \cdot e^{\frac23 - \beta_i}~.
\end{equation}

Traditional methods generally assume that increasing the SNR reduces compression performance while enhancing data analytics accuracy~\cite{wang2022frequency}. However, our experimental results challenge this oversimplified assumption by showing that excessively high SNR is unnecessary for contributing to higher outlier detection accuracy, as illustrated in~\autoref{fig:varingSNR}. 

\subsection{Compressed Data Transformation}

\name employs adaptive base error to control the way we extract semantics. The rationale for the adaptive base error threshold is visualized in~\autoref{fig:fluctuation}. For volatile data, it uses a smaller error bound automatically, which facilitates the retention of data features in the captured semantics and the detection of significant changes. Conversely, for a data series with low variability, it opts for a larger error bound, as minor fluctuations are often not relevant to the overall analysis and may only represent noise. A larger error bound thus allows for more effective compression while retaining more data points. We design the compressed data transformation based on the new paradigm of the adaptive shrinking cone. To allow outlier detection on \emph{Base}, {we propose two heuristic methods, Segment Filter and Base Filter, for transforming compressed data into suitable representations for analytics.

\begin{figure}[!h]
\vspace{-1mm}
\centering
\includegraphics[width=\linewidth]{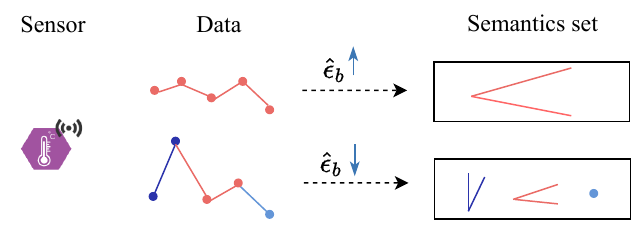}
\vspace{-3mm}
\caption{$\hat{\epsilon}_b$ drops as data fluctuates to preserve data patterns with narrowing span of shrinking cone.}\label{fig:fluctuation}
\vspace{-1mm}
\end{figure}

\noindent \textbf{1) Segment Filter} preserves the segments containing a minimal number of points, and their subsequent neighbor segments. Outliers often disrupt the formation of a shrinking cone, resulting in a very short segment. A data point is marked as a candidate outlier in two scenarios: (a) it terminates a segment, but if it is removed, the segment can continue growing, allowing the subsequent points to form a shrinking cone; (b) it initiates a short segment containing fewer than the minimum required number of points (e.g., 3 points).
    
\noindent \textbf{2) Base Filter} preserves the sub-base if it contains fewer cones than the defined minimum required number, (e.g., 5 merged ones). For sequence-based outliers, the aforementioned principle alone is not sufficient for efficient detection. These sequence outliers exhibit unusual patterns across the dataset, indicating fewer merging shrinking cones within the \emph{Base}. They can be identified by their timestamp tags, which are easily found in the list of corresponding timestamps. For the remaining sub-bases, we retain the representation with the longest segment and discard the others within the same sub-base. This principle is based on the observation that the longest segment representation contains sufficient information relevant for outlier detection.

The Segment Filter method is detailed in Algorithm~\ref{alg:segmentfilter}. This algorithm takes the \emph{Base} $B$, index $i$ and a threshold $\epsilon^*$ as inputs. The algorithm initiates by extracting segments $S$ from $B$ and defining an empty set $P$ to store the key representative segments, as indicated in lines~\ref{PF:line:1}--\ref{PF:line:2}. Once the end of the segment list, $\lvert S \rvert$, is reached, the function terminates, following the steps in lines~\ref{PF:line:3}--\ref{PF:line:4}. For each segment, if the number of points is less than or equal to $\epsilon^*$, the segment is identified as a candidate for outlier detection. In this case, both the current segment and its forwarding neighbor segment are added to $P$ for potential further analysis (lines~\ref{PF:line:5}--\ref{PF:line:9}). If the following segment exceeds the point threshold $\epsilon^*$, it is skipped. The recursive structure of the function ensures that all relevant segments, up to the specified threshold, are systematically processed until the end of the list, see line~\ref{PF:line:10}.

\begin{algorithm}[ht]
\label{alg:segmentfilter}
\caption{SegmentFilter($B$, $i$, $\epsilon^*$)}
\textbf{Inputs:} $B$ - Base, $i$ - index of segment, $\epsilon^*$ - segment threshold  \\
\textbf{Output:} $P$ - Data pattern set \\
    $P \leftarrow \emptyset$\;                                      \label{PF:line:1}
    $S \leftarrow \text{load segments from compressed Base } B$\;      \label{PF:line:2}
    \If{$i \geq \lvert S \rvert$}{                                  \label{PF:line:3}
        \Return\;                                                    \label{PF:line:4}
    }
    \If{$S[i].\text{length} \leq \epsilon^*$}{    \label{PF:line:5}
        $P \leftarrow P \cup \{ S[i] \}$\   \tcp*{Add segment i}                              \label{PF:line:6}
        \If{$S[i+1].\text{length} \geq \epsilon^*$}{ \tcp{Contain sufficient points}               \label{PF:line:7}
            $i \leftarrow i + 1$\;                                       \label{PF:line:8}
            $P \leftarrow P \cup \{ S[i] \}$\     \tcp*{Add segment i+1}                      \label{PF:line:9}
        }
    }
    \Return{SegmentFilter($B$, $i+1$, $\epsilon^*$)}  \tcp*{Otherwise}                   \label{PF:line:10}
\end{algorithm}


\begin{algorithm}
\label{alg:basefilter}
\caption{BaseFilter($S$, $P$, $\epsilon^*$)}
\textbf{Input:} $S$ - Segments, $P$ - Data pattern set, $\epsilon^*$ - Base threshold  \\
\textbf{Output:} $P$ - Data pattern set \\
    $S \leftarrow S - P$\;                                            \label{SF:line:1}
    $S.\text{groupby}(key=(\Theta, K))$\;                             \label{SF:line:1.1}
    \ForEach{$s \in S$}{                                                \label{SF:line:2}
        \If{$s[t].\text{length} \leq \epsilon^*$}{                      \label{SF:line:3}
            $P \leftarrow P \cup \{ s \}$\;                              \label{SF:line:4}
        }
        \Else{                                                             \label{SF:line:5}
            $P \leftarrow P \cup \{ \text{max\_seg\_len}(s) \}$\;          \label{SF:line:6}
        }
    }
    \Return{$P$}\;                                                         \label{SF:line:7}
\end{algorithm}

The Base Filter method is detailed in Algorithm~\ref{alg:basefilter}, which takes the segments set $S$, the current data semantics set $P$, and a threshold $\epsilon^*$ as inputs. The algorithm first removes any previously selected semantics from $S$, updating $S$ to exclude $P$ in line~\ref{SF:line:1}. Then, it groups the segments by their respective origin point value $\Theta$ and derived slope value $K$ of the linear segment to facilitate the filtering process in line~\ref{SF:line:1.1}. The algorithm iterates over each segment $s$ in the updated $S$ (line~\ref{SF:line:2}). For each segment, if its length is less than or equal to $\epsilon^*$, the segment is added to the semantics set $P$ (line~\ref{SF:line:3}--\ref{SF:line:4}). If the number of merged cones within one sub-base exceeds the threshold, the longest segment within that grouping is selected and added to $P$ instead (line~\ref{SF:line:5}--\ref{SF:line:6}). 
The function concludes by returning the updated semantics set $P$ (line~\ref{SF:line:7}). This approach ensures that the final representation captures the longest and most relevant segments necessary for outlier detection, thus streamlining the data for effective analysis.

Both algorithms emphasize the most meaningful data and effectively discard less informative segments. Note that it is important to run Algorithm~\ref{alg:segmentfilter} first in semantic decoding. If not, Algorithm~\ref{alg:basefilter} might filter out neighboring data points that are crucial for accurate outlier detection, particularly when outliers are close to these segments. As shown in~\autoref{fig:outliertype}, if the neighboring segment of point outlier is removed by the BaseFilter algorithm, there is a high chance that the outlier could be misclassified as normal, leading to detection errors.

\section{Evaluation}

In this section, we present the experimental results, comparing the performance of our outlier detection directly on compressed data to that on uncompressed data. The datasets~\footnote{Datasets can be gained from https://github.com/xkitsios/sim-piece} FaceFour (FA), MoteStrain (MS) and ECG (EC) from the UCR time series repository~\cite{dau2019ucr} are used. For these three datasets, we randomly introduce~100 outliers to simulate abnormal behaviors, emulating real-world scenarios where unexpected anomalies occur. Additionally, we incorporate three publicly available outlier detection datasets from the KDD21 competition~\cite{keogh2021multi}, a comprehensive collection of 250 time series with labeled outliers, specifically released as part of the SIGKDD 2021 competition. Among the KDD21\footnote{Datasets can be gained from https://github.com/TheDatumOrg/TSB-UAD} datasets, we use DISTORTEDECG (DE), DISTORTED2sddb (DS), and AirTemperature (AT). \pk{These datasets contain labeled outliers as follows: DE has~300 outliers among 199\,700 data points, DS has~300 outliers among 79\,701 data points, and AT has~1 outlier among 8\,183 data points.} As these datasets come as labeled files, we do a simple processing to convert them into normal time series data before compression.


\begin{figure}[ht]
\vspace{-1mm}
\centering
\includegraphics[width=\linewidth]{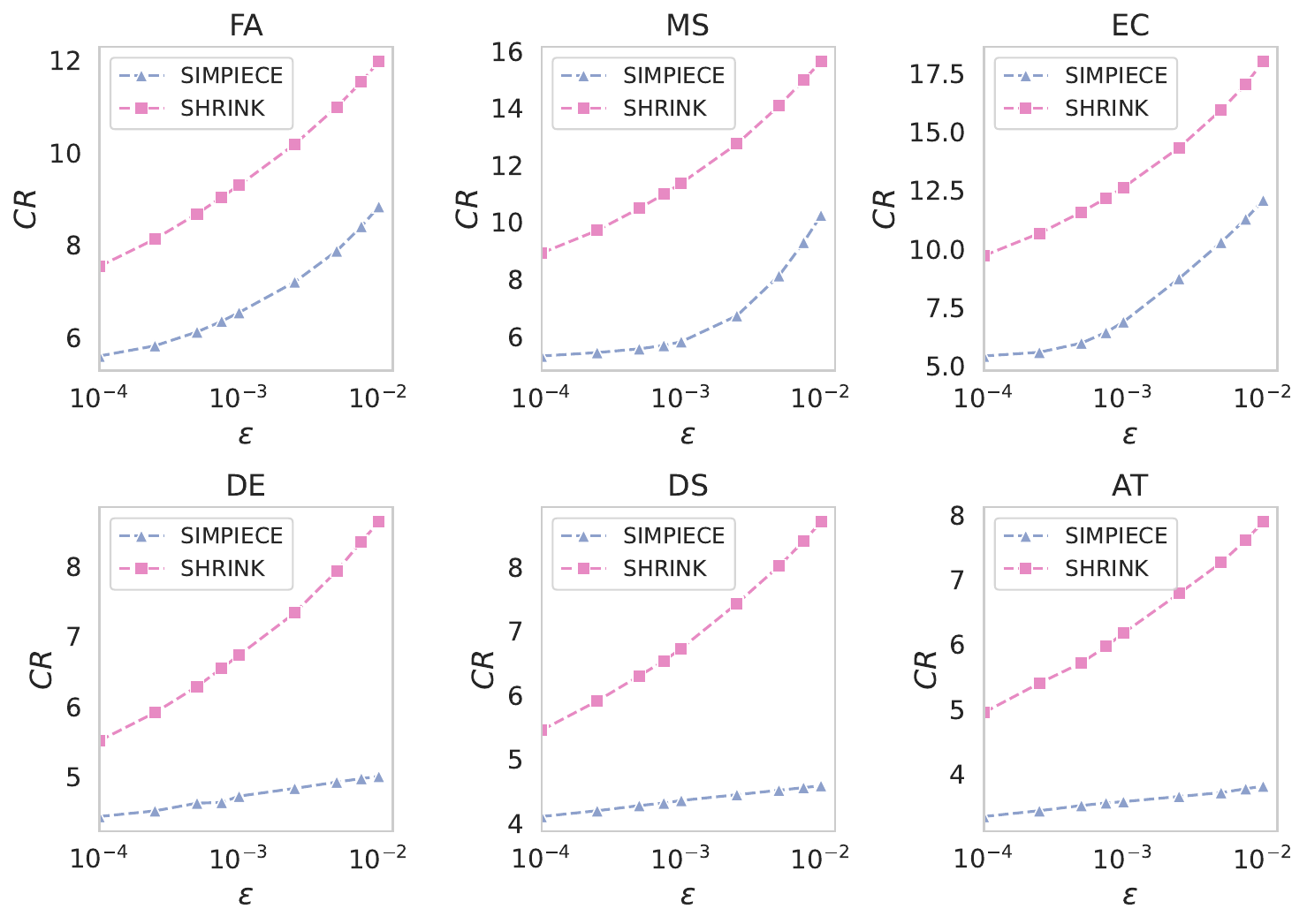}
\vspace{-3mm}
\caption{Comparison of compression ratio; higher CR better compression.}\label{fig:cr}
\vspace{-1mm}
\end{figure}

\subsection{Compression}

As illustrated in \gy{Section~\ref{sec:SemQuant}}, we first need to determine the semantic quantization level so as to ensure the quality of direct data analytics. We set SNR $\eta=25$~dB as this value is commonly used in applications where moderate fidelity is sufficient and storage efficiency is essential~\cite{teh2020sensor}. \gyz{\autoref{fig:cr} shows the compression performance between semantic-aware compression $\name$ and the recently proposed competitive PLA method  \gy{\simpiece~\cite{kitsios2023sim}} over different datasets.} It shows that at any given max error bound $\name$ achieves a better compression ratio, because it can effectively capture the semantics feature and encodes the residual compactly. For direct data analytics, only the compressed \emph{Base} is utilized.

\begin{figure}[!h]
\vspace{-1mm}
\centering
\includegraphics[width=0.9\linewidth]{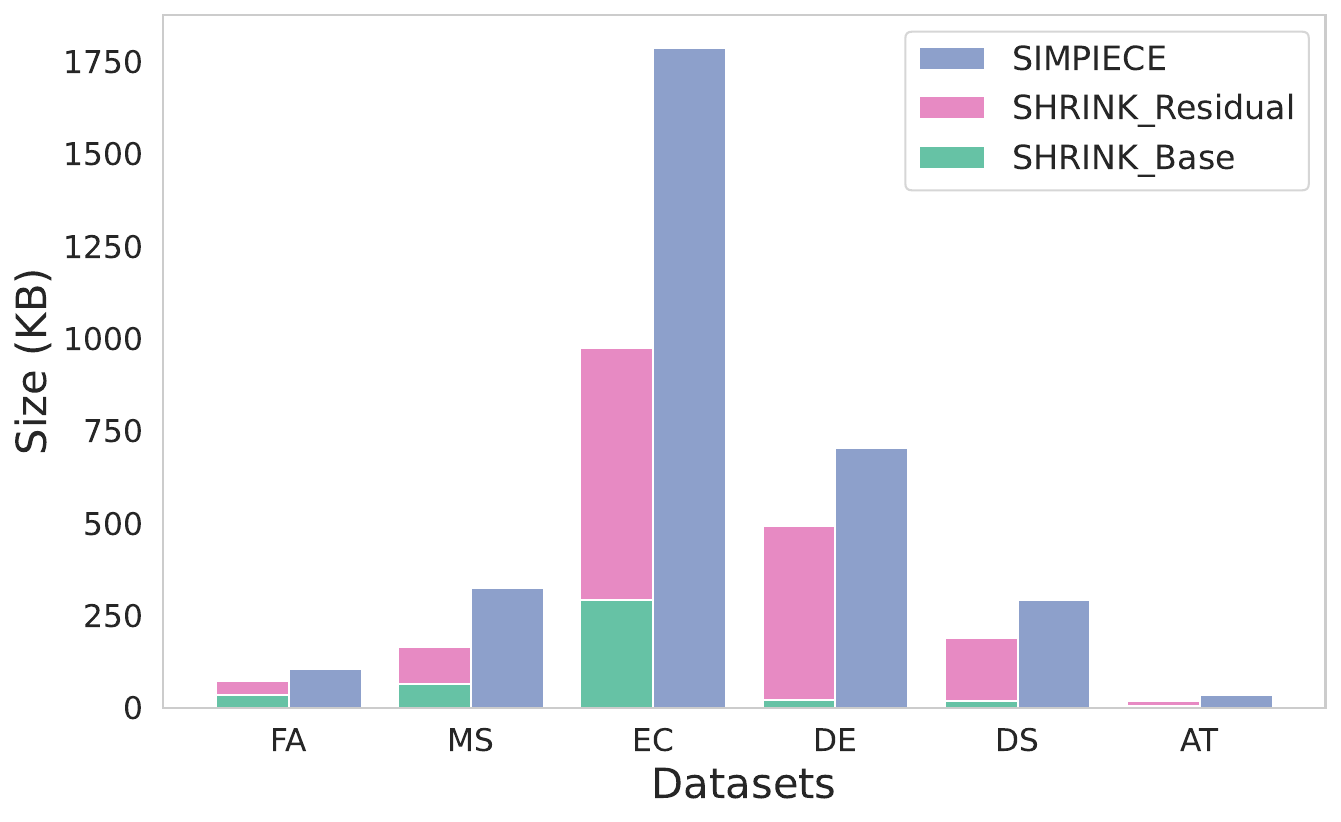}
\vspace{-2mm}
\caption{Composition of compressed data sizes across datasets.}\label{fig:compressSize}
\vspace{-1mm}
\end{figure}

\pk{Since \name organizes compressed data into \emph{Base} and \emph{Residual}, it is worth assessing the \emph{Base} size compared to the total compressed data size. \autoref{fig:compressSize} breaks down the compressed data size for \name and \simpiece with error threshold~$\epsilon=0.001$. Notably, the \emph{Base} which represents the underlying data semantics, remains relatively small in size, around 5\%{-}10\% for the original data size. By contrast, the residuals correspond to a large fraction of the compressed data size. Direct data analytics using \emph{Base} alone will facilitate edge analytics by storing a small Base at the edge.}

\subsection{Outlier Detection Accuracy}

We evaluate performance of outlier detection on compressed data with IForest,  DBScan and AutoEncoder across six datasets using ROC AUC and PR AUC metrics. \pk{DBScan requires the MinPts and~$\epsilon$ parameters; we choose~$\epsilon \in [0.1, 1]$ and MinPts $\in [3, 15]$ via grid search. We set the number of outliers for IForest and AutoEncoder as in the ground truth.} To render results comparable, we introduce the \emph{Detection Ratio} (DR), which normalizes ROC (or PR) values relative to their uncompressed data counterparts, set at 100\%. A DR of 100\% indicates no change in detection accuracy compared to uncompressed data, while DR $>$ 100\% indicates improved detection accuracy on compressed data. To ensure fair evaluation, outliers identified in the compressed data are mapped back to their corresponding labels in the uncompressed data. Thereby, we consistently assess how well the compressed data preserves critical anomaly features aligned with the uncompressed data.


\autoref{fig:outlierMetrics} reports the DR across all datasets. Generally, direct analytics on compressed data achieves comparable outlier detection accuracy, with only a 1\% reduction in the worst case. In multiple cases, direct analytics on compressed data outperforms the baselines, such as performing IForest on compressed data of EC, DE, DS and AT datasets. This suggests that data compression effectively reduces noise and amplifies relevant semantics, thus contributing to anomaly detection. DBScan and AutoEncoder on compressed data maintain performance close to that of the baseline, demonstrating the effectiveness of direct outlier detection on \name compressed data.

\begin{figure}[ht]
\centering
\includegraphics[width=\linewidth]{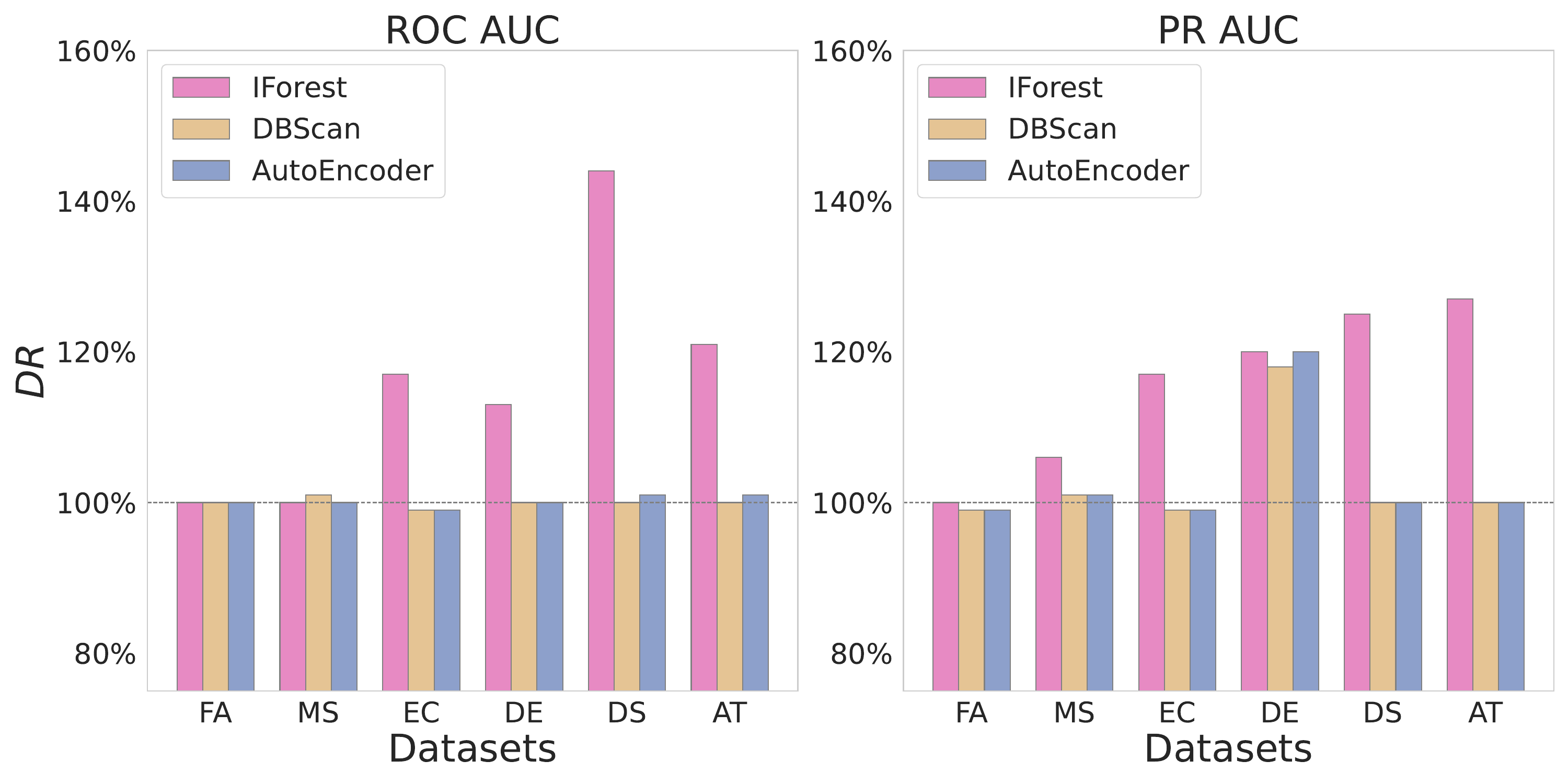}
\vspace{-2mm}
\caption{Outlier detection results. All values are expressed in percentage relative to results with uncompressed data, which corresponds to 100\%.}\label{fig:outlierMetrics}
\vspace{-1mm}
\end{figure}
\begin{table}[!ht]
\centering
\vspace{-1mm}
\caption{Runtime (in Seconds) Comparison Across Approaches}\label{tab:Time}
\begin{tabular}{l c c c c c c}
\toprule
\textbf{} & \textbf{FA} & \textbf{MS} & \textbf{EC} & \textbf{DE} & \textbf{DS} & \textbf{AT} \\ \midrule
\textbf{IForest} & 0.36 & 0.72 & 4.35 & 1.15 & 0.53 & 0.16 \\
                 & \textbf{0.17} & \textbf{0.22} & \textbf{0.69} & \textbf{0.45} & \textbf{0.19} & \textbf{0.13} \\ \midrule
\textbf{DBScan}  & 3.30 & 40.59 & 44.21 & 0.90 & 2.79 & 0.06 \\
                 & \textbf{0.23} & \textbf{3.56} & \textbf{0.92} & \textbf{0.40 }& \textbf{0.13} & \textbf{0.02} \\ \midrule
\textbf{AutoEncoder} & 10.76 & 26.04  & 60.28 & 53.59 & 88.97 & 9.93 \\
                     & \textbf{4.86} & \textbf{11.23} & \textbf{10.03} & \textbf{25.08} & \textbf{21.1} & \textbf{4.92} \\ \bottomrule
\end{tabular}
\vspace{-1mm}
\end{table}

\subsection{Outlier Detection Runtime}

\autoref{tab:Time} outlines the runtime for each detection method. The runtime in bold font is that running on compressed data. IForest consistently exhibits the shortest runtime across all datasets. AutoEncoder, using a neural network with modest parameters, in general, requires the longest runtime, especially on uncompressed data. The runtime of DBScan varies a lot across different datasets, and performing it on compressed data achieves \gy{significant} runtime speedup, on average $16 \times$ and even up to $48 \times$. \gy{The significant improvement can be attributed to its worst-case quadratic runtime of $O(n^2)$~\cite{gan2015dbscan}.} DBScan computes the neighborhood graph of a dataset and uses the connected components of high-degree nodes to determine the clusters. When the neighborhood graph is too large, it becomes costly to compute. For datasets like EC, which contains approximately 700,000 data points, and MS, which is highly dense, their neighborhood graphs become quite large. In these cases, DBScan benefits significantly from our method.

\pk{Although IForest demonstrates the best runtime, it relies on assumptions regarding the percentage of outliers in the data, the same in AutoEncoder. In contrast, DBScan makes few assumptions about the data, determining the number of outliers automatically and handling data of arbitrary shape. As modern datasets continue to grow in size and complexity, unsupervised procedures like DBScan are becoming increasingly important. However, the drawback of worst-case quadratic runtime has limited its usage in large datasets. Our method alleviates these constraints, enabling faster clustering with comparable precision, making it more valuable for large-scale data analytics.}

\subsection{Varying SNR}

As \pk{the precision of data points varies across datasets, it takes considerable effort to adjust the~$\epsilon_b$ parameter for each dataset when using \name. To address this concern, we perform semantic quantization using SNR as a control measure to standardize and simplify the process. By adjusting SNR, we aim to quantify the effect of noise on both compression performance and outlier detection, making it more applicable across datasets without the need for dataset-specific parameter tuning. To illustrate the impact of SNR, we measure the compression and outlier detection performance at an error threshold $\epsilon=0.001$, using the MS dataset and IForest.}

\begin{figure}[!h]
\vspace{-1mm}
\centering
\includegraphics[width=\linewidth]{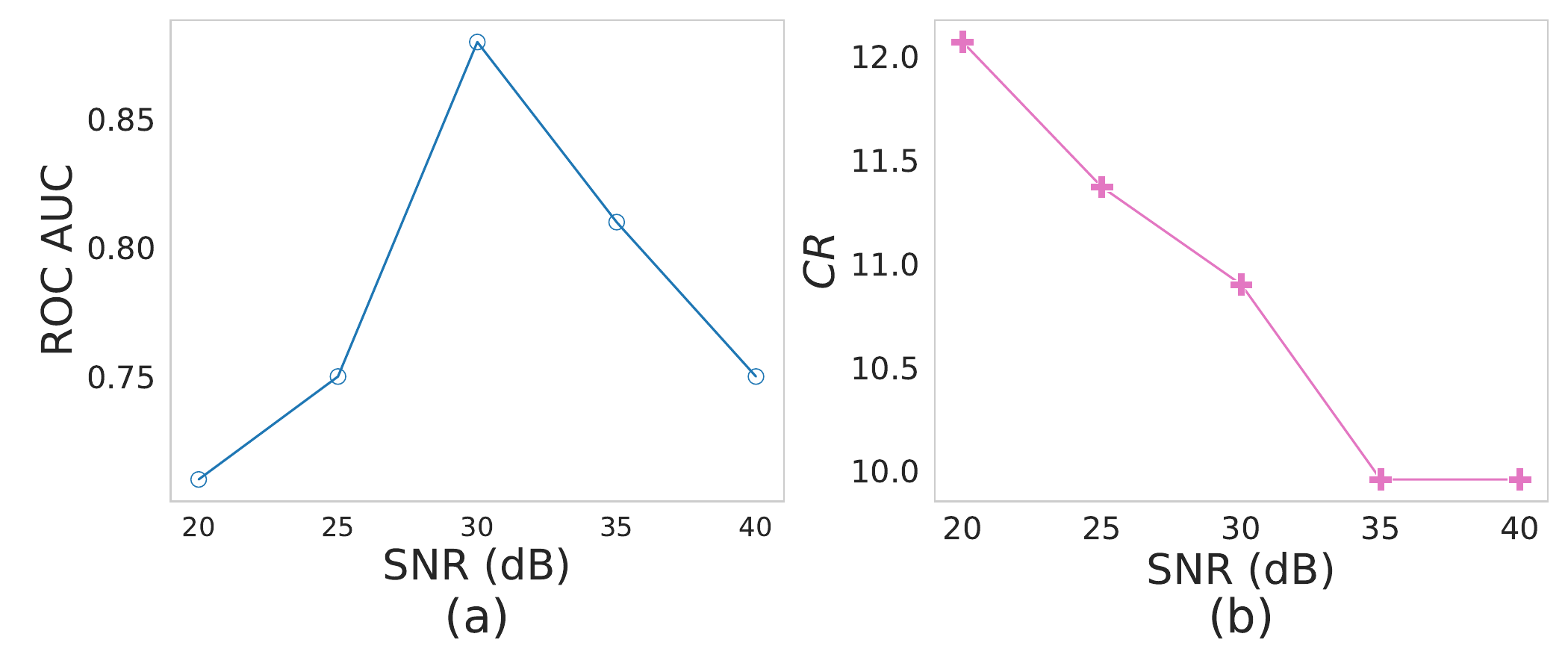}
\vspace{-3mm}
\caption{Evaluation on varing SNR.}\label{fig:varingSNR}
\vspace{-1mm}
\end{figure}

As shown in~\autoref{fig:varingSNR} (a), the ROC AUC score remains high when SNR between of 25 dB and 35 dB, indicating robust outlier detection performance with this SNR range. For SNR $<$ 25 dB, the compressed data experiences excessive information loss, which negatively impacts the detection accuracy. This is due to insufficient preservation of essential semantics in \emph{Base}, making it challenging to distinguish between normal and anomalous data points. For SNR $>$ 35 dB, while more information is retained in the compressed data, this also results in the inclusion of significant noise. Retaining large amounts of noise from the original data may obscure subtle outliers, thereby reducing detection accuracy. From a compression perspective, as shown in~\autoref{fig:varingSNR} (b), as SNR increases, the compression gradually deteriorates. In general, higher SNR values prioritize retaining more data detail, which reduces the achievable compression rate. However, the decline in compression reaches plateau beyond a certain SNR value, such as 35 dB in this example. By focusing on preserving essential semantics, we achieve both reliable outlier detection and effective compression.
\section{Conclusion}


\pk{We introduced an efficient approach for direct analytics on compressed data using \name. By incorporating quantization in semantic encoding and data transformation in semantic decoding, our framework extends \name to task-oriented analytics directly on compressed data, aligning with a semantic communication framework. Using outlier detection as a case study, we have shown that our method achieves detection accuracy comparable to that on uncompressed data, with a drop of less than~1\% in the worst case. Moreover, it accesses approximately~10\% of the data size and accelerates analytics by a factor of four on average. These results establish that our method handles analytics tasks like outlier detection while significantly reducing data transmission and storage costs, emerging as a valuable for resource-constrained IoT systems.}

\section*{Acknowledgment}


\gyz{This work is supported by Independent Research Fund Denmark Light-IoT project \emph{Analytics Straight on Compressed IoT Data} (Grant No. 0136-00376B),  Innovation Fund Denmark GreenCOM project (Grant No. 2079-00040B), NordForsk Nordic University Cooperation on Edge Intelligence (Grant No. 168043) and Aarhus University DIGIT Centre.}
\bibliographystyle{unsrt}
\bibliography{sample}

\end{document}